\pdfoutput=1

\documentclass[11pt]{article}

\usepackage[]{acl}
\usepackage{todonotes} 
\usepackage{amsmath}
\usepackage{cleveref}

\usepackage{tcolorbox}
\usepackage{lipsum}
\usepackage{times} 
\usepackage{geometry}
\usepackage{enumitem} 

\usepackage{multirow}
\usepackage{amssymb}
\usepackage{pifont}
\usepackage{xspace}
\newcommand{\cmark}{\ding{51}}%
\newcommand{\xmark}{\ding{55}}%
\newcommand*{\yoruba}{Yor\`ub\'a\xspace}

\definecolor{bblue}{HTML}{4F81BD}
\definecolor{rred}{HTML}{C0504D}
\definecolor{ggreen}{HTML}{9BBB59}
\usepackage{times}
\usepackage{latexsym}

\usepackage{pgfplots}
\usepackage{pgfplotstable}
\usepackage{booktabs}
\usepackage{xspace}
\pgfplotsset{compat=1.17}

\usepackage[T4,T1]{fontenc}
\usepackage[verbose]{newunicodechar}

\usepackage[utf8]{inputenc}
\usepackage{csquotes}

\usepackage{microtype}

\usepackage{inconsolata}

\usepackage{graphicx}
\usepackage{booktabs}
\usepackage{makecell}

\usepackage{colortbl}

\newcommand*{\mto}{mT0-small\xspace}
\newcommand*{\aya}{Aya-23-35B\xspace}
\newcommand*{\bloom}{BLOOMZ 7B \xspace}
\newcommand*{\gemma}{Gemma 2 9B\xspace}
\newcommand*{\gemmaTwoL}{Gemma 2 27B\xspace}

\newcommand*{\llamal}{LLaMa 3.1 70B\xspace}
\newcommand*{\mistral}{Mistral 7B\xspace}
\newcommand*{\inkuba}{InkubaLM-0.4B\xspace}
\newcommand*{\gpto}{GPT-4o\xspace}

\title{AfriHate: A Multilingual Collection of Hate Speech and Abusive Language Datasets for African Languages}

\author{
\textbf{Shamsuddeen Hassan Muhammad}$^{1,2}$\thanks{Equal contribution}, \textbf{Idris Abdulmumin}$^{3*}$, \textbf{Abinew Ali Ayele}$^{4,5}$,\\ \textbf{David Ifeoluwa Adelani}$^{6}$,
\textbf{Ibrahim Said Ahmad}$^{2,7}$, \textbf{Saminu Mohammad Aliyu}$^{2}$, \\ \textbf{Nelson Odhiambo Onyango}$^{8}$, 
\textbf{Lilian D. A. Wanzare}$^{8}$, \textbf{Samuel Rutunda}$^{9}$,\\ \textbf{Lukman Jibril Aliyu}$^{10}$, \textbf{Esubalew Alemneh}$^{11}$, \textbf{Oumaima Hourrane}$^{12}$,\\
\textbf{Hagos Tesfahun Gebremichael}$^{4}$, \textbf{Elyas Abdi Ismail}$^{11}$, \textbf{Meriem Beloucif}$^{13}$,\\ \textbf{Ebrahim Chekol Jibril}$^{14}$,
\textbf{Andiswa Bukula}$^{15}$, \textbf{Rooweither Mabuya}$^{15}$, \textbf{Salomey Osei}$^{16}$,\\ \textbf{Abigail Oppong}$^{17}$, \textbf{Tadesse Destaw Belay}$^{18,19}$,
\textbf{Tadesse Kebede Guge}$^{20}$,\\ \textbf{Tesfa Tegegne Asfaw}$^{4}$, \textbf{Chiamaka Ijeoma Chukwuneke}$^{21}$,
\textbf{Paul Röttger}$^{22}$,\\ \textbf{Seid Muhie Yimam}$^{5}$, \textbf{Nedjma Ousidhoum}$^{23}$ \\[1mm]
\footnotesize $^{1}$Imperial College London, $^{2}$Bayero University Kano, $^{3}$DSFSI, University of Pretoria, $^{4}$Bahir Dar University,\\
\footnotesize $^{5}$University of Hamburg, $^{6}$Mila, McGill University \& Canada CIFAR AI Chair, $^{7}$Northeastern University, $^{8}$Maseno University, \\
\footnotesize $^{9}$Digital Umuganda, $^{10}$HausaNLP, $^{11}$Haramaya University, $^{12}$Al Akhawayn University, $^{13}$Uppsala University,\\
\footnotesize $^{14}$Istanbul Technical University, $^{15}$SADiLaR, $^{16}$University of Deusto, 
 $^{17}$Independent Researcher, $^{18}$Instituto Politécnico Nacional,\\
\footnotesize $^{19}$Wollo University, $^{20}$Addis Ababa University, $^{21}$Lancaster University, $^{22}$Bocconi University, $^{23}$Cardiff University\\
\footnotesize \texttt{Contact: s.muhammad@imperial.ac.uk, seid.muhie.yimam@uni-hamburg.de}
}

\begin{document}
\maketitle
\begin{abstract}

Hate speech and abusive language are global phenomena that need socio-cultural background knowledge to be understood, identified, and moderated. However, in many regions of the Global South, there have been several documented occurrences of (1)\ absence of moderation and (2)\ censorship due to the reliance on keyword spotting out of context. Further, high-profile individuals have frequently been at the center of the moderation process, while large and targeted hate speech campaigns against minorities have been overlooked.
These limitations are mainly due to the lack of high-quality data in the local languages and the failure to include local communities in the collection, annotation, and moderation processes. To address this issue, we present AfriHate: a multilingual collection of hate speech and abusive language datasets in 15 African languages. Each instance in AfriHate is annotated by native speakers familiar with the local culture. We report the challenges related to the construction of the datasets  various classification baseline results with and without using LLMs.\footnote{The datasets, individual annotations, and hate speech and offensive language lexicons are available on \url{https://github.com/AfriHate/AfriHate}}

\textcolor{red}{\textbf{Content Warning:} This paper contains representative examples of hate speech and offensive language.}

\end{abstract}

\section{Introduction}

\begin{displayquote}
    \textit{No one is born hating another person because of the color of his skin, or his background, or his religion. People must learn to hate, and if they can learn to hate, they can be taught to love, for love comes more naturally to the human heart than its opposite.} -- \cite{mandela1994long}
\end{displayquote}

\begin{figure*}[!htb]
    \centering
        \includegraphics[clip, trim=0.4cm 17.7cm 0.6cm 0.7cm, width=0.99\textwidth]{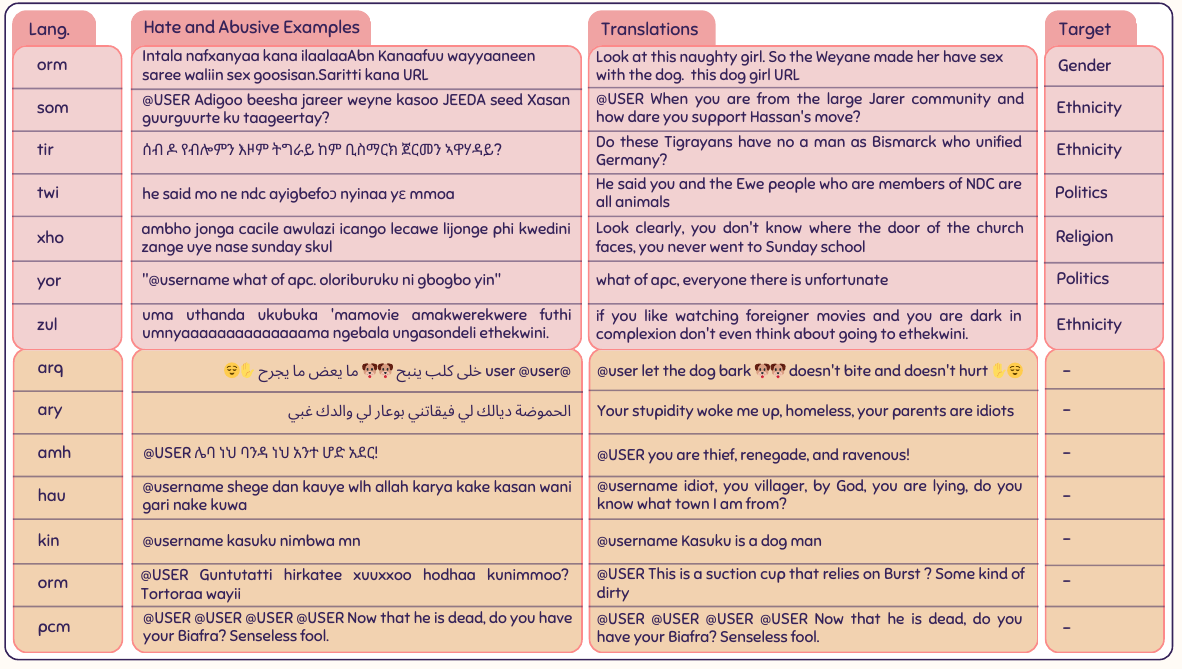}
     \vspace*{-3mm}
     \captionof{table}[]{\textbf{Examples of hateful and abusive instances in AfriHate}. All  hateful posts are assigned targets. } 
    \label{fig:Afrihate_examples}
\end{figure*}

Hate speech and abusive language are global phenomena that highly depend on specific socio-cultural contexts. Although hate speech and abusive language deviate from the norm on social media, they quickly attract significant attention, spread among online communities \cite{mathew2019spread}, and incite harm or violence on individuals in real life \cite{Saha2019-wl}. Tangible efforts to address these problems must take social and cultural contexts into account \cite{shahid2023decolonizing}. However, in the absence of high-quality data or when excluding local voices from the collection and annotation processes, one may fail to build assistive tools that help address the problem and moderate such content.

Collecting hate speech and offensive language datasets is complex and time-consuming as researchers typically rely on keywords, hashtags, or user accounts to build datasets \cite{ousidhoum2020comparative}. They may need further insights from both moderators \cite{arora2023detecting} and affected communities \cite{maronikolakis2022listening}.
Further, resources in languages other than English are scarce, especially for low-resource languages. 

To bridge the current gap in the area, we present \textbf{\textsc{AfriHate}}
a collection of new hate speech and abusive language datasets in 15 languages spoken in various African regions: Algerian Arabic, Amharic, Igbo, Kinyarwanda, Hausa, Moroccan Arabic, Nigerian Pidgin, Oromo, Somali, Swahili, Tigrinya, Twi, isiXhosa, \yoruba, and isiZulu. The datasets are annotated by native speakers and include three classes:
\textbf{hate}, \textbf{abusive/offensive}, or \textbf{neutral}--neither hateful nor abusive. The targets of the hateful tweets were further labeled based on six common attributes used to discriminate against people: \textbf{ethnicity}, \textbf{politics}, \textbf{gender}, \textbf{disability}, \textbf{religion}, or \textbf{other}. \Cref{fig:Afrihate_examples} shows a sample of the datasets in various languages.

We report the data collection and annotation strategies and challenges when building AfriHate, present various classification baselines with and without using LLMs, and discuss the results. We find that the performance highly depends on the language and that multilingual models can help us boost the performance in low-resource settings. 
We publicly release the datasets, and individual labels, in addition to manually curated hate speech and offensive language lexicons. These provide a valuable foundation for the research community interested in hate speech and abusive language, African languages, and researchers interested in studying disagreements. 

\section{Related Work} 

\begin{table*}[!htb]
    \centering
    \resizebox{\textwidth}{!}{
        \begin{tabular}{@{}llllll@{}}
        \toprule[1.5pt]
        
        \textbf{Language} & \textbf{Code} & \textbf{Subregion} & \textbf{Spoken in} 
        & \textbf{Script}\\
            \midrule

          Algerian Arabic/Darja & \texttt{arq}   & North Africa  & Algeria  
        & Arabic \\
        Amharic & \texttt{amh}   & East Africa   & Ethiopia, Eritrea  
        & Ethiopic\\
      
         Hausa  & \texttt{hau}   & West Africa  & Northern Nigeria, Niger, Ghana, and Cameroon,  
         & Latin\\
        Igbo  & \texttt{ibo}   & West Africa    & Southeastern Nigeria  
        & Latin\\
        Kinyarwanda  & \texttt{kin}  & East Africa  & Rwanda    
        & Latin \\
        Moroccan Arabic/Darija  & \texttt{ary}  & North Africa  & Morocco   
        & Arabic/Latin \\
        Nigerian Pidgin  & \texttt{pcm}   & West Africa  & Nigeria, Ghana, Cameroon,  
        & Latin\\
        Oromo  & \texttt{orm} & East Africa & Ethiopia, Kenya, Somalia 
        & Latin \\
        Somali  & \texttt{som} & East Africa & Somalia, Ethiopia, Djibouti, Kenya 
        & Latin\\ 
        Swahili  & \texttt{swa} & East Africa & Kenya, Tanzania, Uganda, DR Congo, Rwanda, Burundi, Mozambique & Latin \\
        Tigrinya  & \texttt{tir}  & East Africa    & Ethiopia, Eritrea  
        & Ethiopic\\
        
        Twi   & \texttt{twi}   & West Africa   & Ghana 
        & Latin\\
        
        Xhosa  & \texttt{tso}   & Southern Africa   & Mozambique, South Africa, Zimbabwe, Eswatini 
        & Latin\\
         \yoruba  & \texttt{yor}  & West Africa   & Southwestern and Central Nigeria, Benin, and Togo  
         & Latin\\
          Zulu  & \texttt{zul}  & Southern Afric   & Southern Africa 
         & Latin\\
        \bottomrule[1.5pt] 
        \end{tabular}
    }
    \caption{\textbf{Information about the AfriHate languages}: their ISO codes, subregions, countries in which they are mainly spoken, and the writing scripts included in AfriHate.}
    \label{tab:languages}
    \vspace*{6mm}
\end{table*}

The fast-spreading nature of hate speech and abusive language have been at the center of a significant amount of NLP work in recent years \cite{Talat2016HatefulSO,Vigna2017HateMH,basile2019semeval,Mansur2023TwitterHS}. 
However, as there is no unanimous definition of hate speech, researchers in the area have adopted different ones when building resources. For instance, some studies define hate speech as any speech that can cause danger or harm to disadvantaged groups \cite{davidson2017automated}, others focus on whether the speech is intended to promote hatred \cite{gitari2015lexicon}, or whether it dehumanises protected groups \cite{vidgen-etal-2021-learning}, which leads to various challenges such as the lack of generalisability \cite{yin2021towards}.

Despite Africa being home to more than 2,000 languages and the increasing interest in building hate speech and offensive languages resources for non-English languages \cite{ousidhoum2019multilingual,Madeddu2023DisaggregHateIC,rottger-etal-2022-multilingual}, few datasets focus on African languages, such as Amharic \cite{Ayele2024ExploringBA,ayele20225js}, Afaan Oromo \cite{ababu2022afaan}, \yoruba \cite{ilevbare2024ekohate}, Hausa \cite{vargas2024hausahate,adam2023detection}, and Nigerian Pidgin \cite{ndabula2023detection,ilevbare2024ekohate,aliyu2022herdphobia, tonneau-etal-2024-naijahate}. Moreover, most resources adopt a binary labeling scheme (hate/offensive) and do not include a normal class (e.g., \cite{aliyu2022herdphobia}), or do not add the target attributes (e.g., \cite{ilevbare2024ekohate}). Other work \cite{tonneau-etal-2024-naijahate} relies on active learning for annotating some data instances, which is not ideal when labeling hate speech for under-resourced African languages or when focusing on underrepresented cultures \cite{lee-etal-2023-hate, lee-etal-2024-exploring-cross}. Similarly, a limited number of studies involve African communities in the dataset creation process \citep{masakhaner,maronikolakis2022listening, abdulmumin2024correcting}.
 We take a step forward in addressing this problem by developing 15 new datasets for hate and offensive speech in various languages spoken across the African continent.

\begin{table*}[hbt!]
\centering
\small
\resizebox{\textwidth}{!}{
\begin{tabular}{lrrrrrrrrrrrrrrr}
\toprule[1.5pt]
\textbf{} & \texttt{\textbf{amh}} &\texttt{\textbf{arq}} & \texttt{\textbf{ary}} & \texttt{\textbf{hau}} & \texttt{\textbf{ibo}} & \texttt{\textbf{kin}} & \texttt{\textbf{oro}} & \texttt{\textbf{pcm}} & \texttt{\textbf{som}} & \texttt{\textbf{swa}} & \texttt{\textbf{tir}} & \texttt{\textbf{twi}} & \texttt{\textbf{xho}} & \texttt{\textbf{yor}} & \texttt{\textbf{zul}}\\
\midrule
\textbf{Hate}& 88 & 62 & 41 & 36 & 46 & 264 & 126 & 23 & 45 & 24 & 58 & 20 & 42 & 68 & 31 \\
\textbf{Abusive}& 74 & 40 & 0 & 149 & 118 & 362 & 159 & 26 & 67 & 12 & 66 & 86 & 177 & 109 & 118 \\
\bottomrule[1.5pt]
\end{tabular}
}
\caption{\textbf{Number of keywords used for data collection.} For Algerian  Arabic (arq), the collection includes controversial topics that are neither hateful nor offensive.}

\label{tab:keyword_table}
\end{table*}

\section{Creating AfriHate}

AfriHate covers 15 languages from various African regions. In \Cref{tab:languages}, we report the scripts of these languages and the main regions where they are spoken. The collection includes tweets from 2012 to 2023 collected using the Academic API before the suspension of free academic access. The API allowed us to collect up to 10 million tweets per month, and Twitter/X is a commonly used platform in African countries with documented cases of hate speech propagation \cite{Adjai2013MigrationXA,Egbunike2015NatureOT,Oriola2020EvaluatingML,ridwanullah2024politicization,Raborife2024TheRO}.

\subsection{Data Collection}
Except for Amharic and Tigrinya, the Twitter API does not support African languages, which makes the data collection challenging. We, therefore, follow strategies adopted by previous work such as \citet{muhammad2022naijasenti,muhammad2023afrisenti} and use various heuristics based on hate speech and abusive language keywords, user handles, stopwords, hashtags, and locations. \Cref{tab:keyword_table} shows the number of keywords used for data collection in each language. 

Since interpreting hate and abusive content heavily depends on understanding political and socio-cultural contexts, we have adopted language-dependent collection and annotation strategies. 
As previous work relied on specific keywords to collect data \cite[e.g.,][]{Talat2016HatefulSO,basile2019semeval}, we follow a similar strategy by using a larger set of keywords. Similarly to \citet{ousidhoum2019multilingual}, our keywords include culture-specific controversial topics and are of various sizes (see \Cref{tab:keyword_table}). However, for some languages such as Nigerian-Pidgin and Hausa, an initial pre-annotation phase revealed a limited number of hateful tweets. We therefore used additional strategies to collect more tweets: 1)\ keyword crowdsourcing, 2)\ manual data collection, and 3)\ using existing datasets, as we explain in the following.

\paragraph{Crowdsourcing Keywords}
To crowdsource additional keywords, we first asked native speakers to provide us with a list of hateful, abusive, or controversial keywords. 
To further diversify our lists, we contacted social media influencers, who asked their followers to share abusive or hateful keywords in their local languages by filling in a form created for anonymous collection. This strategy resulted in a broader vocabulary as the followers come from various backgrounds. We also used off-the-shelf curated hate speech lexicons from PeaceTech Lab\footnote{\url{https://www.peacetechlab.org/the-peacetech-toolbox}} and Hatebase.\footnote{\url{https://hatebase.org}} 
The lists were post-processed by native speakers prior to the collection of the tweets.  
\paragraph{Manual Data Collection}

For Kinyarwanda and Twi, native speakers manually collected all the tweets using a combination of keywords and user handles. We curated a list of user handles of public figures who frequently post hateful or abusive content, and collected tweets from their profiles.

\paragraph{Using Existing Datasets}

For Nigerian Pidgin, we used an additional existing hate speech dataset from \citet{tonneau-etal-2024-naijahate}. Since some instances in this dataset were annotated using active learning, we re-annotated those labeled as hateful or offensive only. 
Similarly, for Swahili, we selected instances from the sentiment analysis dataset by \citet{muhammad2023afrisenti}\footnote{We selected only negative instances, as they may contain hateful or abusive content.}, the misinformation dataset by \citet{amol2023politikweli}, and the hate speech one by \citet{ombui2019hate}. 

These instances were re-annotated into our pre-define classes (hate, offensive and normal). We further labeled the targets of the tweets based on our pre-defined target attributes, i.e., disability, ethnicity, gender, politics, religion, and others.

\subsection{Data Processing}

We further cleaned the collected tweets and removed retweets, tweets containing less than three words, duplicates, URLs, invisible characters, and redundant white spaces. We converted the tweets written in Latin script to lowercase and anonymized the tweets by replacing @mentions with a placeholder \textit{@user}.

\subsection{Language Identification}

We collected the tweets using location and keywords. This is particularly challenging for African languages since people within one location can speak different languages. That is, keywords and hashtags may appear in more than one language, which makes the data selection more difficults 

Although ptudies have use \cite{muhammad2022naijasenti, muhammad2023afrisenti} open-source and closed-source language identification (LID) tools, these often show low accuracy in African languages, especially when used in social media posts.
This is largely due to the unique linguistic characteristics of these languages, such as the common usage of code-mixing and digraphia, i.e., a language can be written in more than one script.

To address these limitations, we built a LID model that improves identification performance for social media text data in our target languages through continued pretraining of AfroXLMR model \cite{alabi-etal-2022-adapting} on Glot500-c corpus which covers 511 predominantly low-resource languages \cite{imanigooghari-etal-2023-glot500}. We further fine-tuned the model on AfriSenti-LID dataset, which focuses on social media data and covers most of the languages included in AfriHate and a similar Twitter-sphere. Our LID tool and its documentation can be found on our project page.\footnote{\url{https://github.com/hausanlp/AfriLID}}

\begin{table*}[ht]
\centering
\resizebox{\textwidth}{!}{
    \begin{tabular}{@{}llllllllllllllll@{}}
    \toprule[1.5pt]
     \textbf{Language} & \texttt{\textbf{amh}} &\texttt{\textbf{arq}} & \texttt{\textbf{ary}} & \texttt{\textbf{hau}} & \texttt{\textbf{ibo}} & \texttt{\textbf{kin}} & \texttt{\textbf{oro}} & \texttt{\textbf{pcm}} & \texttt{\textbf{som}} & \texttt{\textbf{swa}} & \texttt{\textbf{tir}} & \texttt{\textbf{twi}} & \texttt{\textbf{xho}} & \texttt{\textbf{yor}} & \texttt{\textbf{zul}} \\ 
    \midrule
    \textbf{Manually Collected} & \xmark & \xmark & \xmark & \cmark & \xmark & \cmark & \xmark & \xmark & \xmark & \xmark & \xmark & \cmark & \xmark & \xmark & \xmark \\
    \textbf{Pre-Annotation} & \xmark & \xmark & \cmark & \cmark & \cmark & \cmark & \xmark & \cmark & \xmark & \xmark & \xmark & \cmark & \cmark & \xmark & \cmark \\ \midrule
    \textbf{Total Annotators} & 11 & 7 & 3 & 3 & 6 & 3 & 9 & 3 & 7 & 5 & 8 & 3 & 3 & 4 & 3 \\
    \textbf{Annotators per Instance} & 4 & 3 & 3 & 3 & 3 & 3 & 4 & 3 & 4 & 3 & 4 & 3 & 3 & 4 & 3 \\\midrule
    \textbf{Free-Marginal Multirater Kappa} $\uparrow$ & 0.63 & 0.68 & 0.61 & 0.75 & 0.80 & 0.81 & 0.63 & 0.65 & 0.46 & 0.55 & 0.46 & 0.75 & 0.62 & 0.68 & 0.81 \\
    \bottomrule[1.5pt]
    \end{tabular}
}
\caption{\textbf{Collection and annotation details} for the AfriHate dataset. The table shows if the data was manually collected, whether a pre-annotation step was conducted, the total number of annotators, the number of annotators per instance, and the inter-annotator agreement (Free-Marginal Multirater Kappa).}
\label{tab:merged_table_strat_IAA}
\end{table*}

\begin{table*}
\centering
\resizebox{\textwidth}{!}{
\begin{tabular}{lrrrrrrrrrrrrrrr}
\toprule
{\textbf{Class}} & \texttt{\textbf{amh}} &\texttt{\textbf{ary}} & \texttt{\textbf{arq}} & \texttt{\textbf{hau}} & \texttt{\textbf{ibo}} & \texttt{\textbf{kin}} & \texttt{\textbf{oro}} & \texttt{\textbf{pcm}} & \texttt{\textbf{som}} & \texttt{\textbf{swa}} & \texttt{\textbf{tir}} & \texttt{\textbf{twi}} & \texttt{\textbf{xho}} & \texttt{\textbf{yor}} & \texttt{\textbf{zul}} 
\\
\midrule
\textbf{Hate} & 2,246 & 162 & 674 & 343 & 251 & 1,268 & 2,293 & 1,177 & 388 & 3,974 & 2,940 & 443 & 210 & 150 & 147 \\
\textbf{Abusive} & 1,353 & 778 & 2,270 & 2,336 & 3,510 & 1,146 & 667 & 5,238 & 1,404 & 7,708 & 1,070 & 3,278 & 1,550 & 2,655 & 1,839 \\
\textbf{Neutral} & 1,359 & 310 & 1,690 & 3,965 & 1,242 & 2,308 & 2,072 & 4,184 & 2,868 & 9,410 & 1,062 & 180 & 1,923 & 2,074 & 2,322 \\
\bottomrule
\end{tabular}}

\caption{\textbf{Number of instances per class in each dataset.}}
\label{tab:instances_per_class}
\end{table*}

\subsection{Data Annotation}

\subsubsection{Pre-Annotation and Data Selection}

We randomly sampled tweets in each language and conducted a pilot annotation, which showed a large class imbalance despite collecting tweets using abusive and hateful keywords. 
For instance, most tweets in Hausa were neutral (neither hateful nor abusive) due to keywords carrying multiple meanings depending on the region where the word is used, i.e., it can have a neutral connotation in some parts of Nigeria. For example, the word \textit{Aboki} means ``friend'' in Northern Nigeria, while it can be an insult in the Southern part of the country.
As this would have led to an insufficient number of instances in each target class, we included a \textbf{pre-annotation} phase to ensure each class covered a reasonable percentage of the data.

During the pre-annotation, we provided the annotators with a distinct pool of tweets and asked them to select those likely to be hateful or abusive. We then aggregated the pre-selected tweets, and multiple annotators labeled them. 
\Cref{tab:merged_table_strat_IAA} indicates the languages for which we conducted a pre-annotation step, i.e., only those for which we observed a significantly high imbalance during the pilot annotation.

\begin{table*}[hbt!]
\centering
\small
\resizebox{\textwidth}{!}{
\begin{tabular}{lrrrrrrrrrrrrrrr}
\toprule
\textbf{Hate Target} & \texttt{\textbf{amh}} &\texttt{\textbf{arq}} & \texttt{\textbf{ary}} & \texttt{\textbf{hau}} & \texttt{\textbf{ibo}} & \texttt{\textbf{kin}} & \texttt{\textbf{oro}} & \texttt{\textbf{pcm}} & \texttt{\textbf{som}} & \texttt{\textbf{swa}} & \texttt{\textbf{tir}} & \texttt{\textbf{twi}} & \texttt{\textbf{xho}} & \texttt{\textbf{yor}} & \texttt{\textbf{zul}} \\ \midrule
\textbf{Disability} & 0 & 0 &26 & 0 & 0 & 6 & 0 & 1 & 0 & 112 & 0 & 32 & 3 & 2 & 0  \\
\textbf{Ethnicity} & 902  & 269 & 150 & 32 &  383 & 94  &  462 & 537 & 74 & 2,799 & 573 & 245 & 94 &   103 & 241  \\
\textbf{Gender} & 15 & 2 &6 & 2 & 3 & 24 & 12 & 50 & 17 & 117 & 0 & 12 & 111 &  13 & 0  \\
\textbf{Politics} & 1,247 & 118 & 63 & 0 & 6 & 1,096 & 1,550 & 87 & 703 & 233 & 2,251 & 32 & 0 & 33 & 0  \\
\textbf{Religion} & 130 & 6 & 207 & 38 & 0 & 10 & 31 & 105 & 19 & 336 & 16 & 14 & 0 & 22 & 0 \\
\textbf{Others} & 199 & 0 & 0 & 0 & 0 & 0 & 767 & 15 & 208 & 501 & 85 & 14 & 0 & 76  & 12  \\

\bottomrule
\end{tabular}
}
\caption{\textbf{Data distribution of hate speech targets in AfriHate}.}
\label{table:data_distribution}
\end{table*}

\begin{table*}[h]
\centering
\resizebox{\textwidth}{!}{
\begin{tabular}{lrrrrrrrrrrrrrrr}
\toprule
\textbf{Split} &  \texttt{\textbf{amh}} &\texttt{\textbf{ary}} &\texttt{\textbf{arq}} & \texttt{\textbf{hau}} & \texttt{\textbf{ibo}} & \texttt{\textbf{kin}} & \texttt{\textbf{oro}} & \texttt{\textbf{pcm}} & \texttt{\textbf{som}} & \texttt{\textbf{swa}} & \texttt{\textbf{tir}} & \texttt{\textbf{twi}} & \texttt{\textbf{xho}} & \texttt{\textbf{yor}} & \texttt{\textbf{zul}} \\ \midrule
\textbf{Train} & 3,467 & 3,240 & 716 & 4,566 & 3,419 & 3,302 & 3,517 & 7,416 & 3,174 & 14,760 & 3,547 & 2,564 & 2,502 & 3,336 & 2,940 \\
\textbf{Dev} & 744 & 695 & 211 & 1,029 & 774 & 706 & 763 & 1,590 & 741 & 3,164 & 760 & 639 & 559 & 724 & 640 \\
\textbf{Test} & 747 & 699 & 323 & 1,049 & 821 & 714 & 759 & 1,593 & 745 & 3,168 & 765 & 698 & 622 & 819 & 728 \\
\bottomrule

\end{tabular}
}
\caption{\textbf{Number of instances included in the training (train), development (dev), and test splits} of the different datasets.}
\label{tab:dataset_split}

\end{table*}

\subsubsection{Recruiting Annotators}
The unavailability of annotators for African languages on common platforms like Amazon Mechanical Turk and Prolific makes traditional crowdsourcing methods impractical. As \citet{kirk2023semeval} demonstrated that trained annotators achieve higher quality results, we trained native speakers and recruited them as annotators. For each language, we also recruited a native speaker as a language lead who would control for the quality of the annotations. We used Label Studio as an annotation platform\footnote{\url{https://labelstud.io/}} and an adapted version of the Potato annotation tool\footnote{\url{https://github.com/davidjurgens/potato}}. 

\subsubsection{Annotation Task}
\paragraph{Labels}
We provided the annotators with thorough guidelines (see \Cref{fig:annotation_guide} in the Appendix). We asked the annotators to choose one of three categories: \textbf{Hate}, \textbf{Abusive/Offensive}, or \textbf{Neutral}. The latter means that the tweet is neither hateful nor abusive. 
Tweets spotted in a language different from the target one were labeled \textbf{Indeterminate} and were later excluded from the final dataset.

Similarly to \citet{Ayele2024ExploringBA,ousidhoum2019multilingual,fortuna2019hierarchically}, annotators had to select the target(s) of the hateful tweets. That is, the common attribute(s) based on which the tweet is discriminating against people: \textbf{Ethnicity}, \textbf{Politics}, \textbf{Gender}, \textbf{Disability}, \textbf{Religion}, or \textbf{Other}. We do not include targets offensive and abusive tweets as these can often be generic and directed towards an individual as previously reported by \citet{zampieri2019predicting} (e.g., see the Algerian Arabic example in Table \ref{fig:Afrihate_examples}).
Details about the targets can be found in \Cref{fig:annotation_guide} in the Appendix. 
For some languages such as Hausa, we asked the annotators to spot which words made them label the tweet hateful or abusive.

\paragraph{Final label selection}
As shown in Table \ref{tab:merged_table_strat_IAA}, for languages where we conducted a pre-annotation step, each tweet was annotated by 3 annotators, leading to a total of 4 labels per tweet with the pre-annotation label counting as one. On the other hand, for instances in datasets for which we did not carry a pre-annotation step, 3 to 4 annotators were assigned to each tweet, and the final gold label was determined by majority voting, i.e., two out of three labels or three out of four labels.

\Cref{tab:instances_per_class} and  \Cref{table:data_distribution} show the final number of instances per class and the target distributions for all the languages. 

\subsubsection{Inter-Annotator Agreement}
To assess the inter-annotator agreement (IAA), we computed the free marginal Randolph's Kappa score \cite{Randolph2005FreeMarginalMK} for each of the datasets. We chose this metric as it is suitable for tasks involving multiple annotators. 

\Cref{tab:merged_table_strat_IAA} shows the IAA scores for all the datasets, the total number of annotators in each, and the number of annotators per instance. The IAA scores range from $0.46$ to $0.81$, indicating medium to high agreement levels. The highest agreement scores are reported for \texttt{kin}, and \texttt{twi}, which can be due to the manual collection of only potentially abusive and hateful tweets. For other languages such as \texttt{hau} and \texttt{zul}, the high agreement can be attributed to the pre-annotation step, which helped us filter tweets that were later annotated

\subsection{Dataset Statistics}
As shown in Table \ref{tab:instances_per_class}, most datasets are imbalanced, and the \textbf{\textit{hate}} class includes fewer instances in 9 out of 15 languages. The variations in the class distributions are due to the differences between the languages and the data collection techniques. Further, the target distributions also differ because of socio-cultural characteristics related to local politics, social dynamics, and an unavoidable degree of selection bias \cite{ousidhoum2020comparative}.


\paragraph{Data Splits}
We split the AfriHate datasets based on the various label distributions. As reported in Table \Cref{tab:dataset_split}, each test set includes a minimum of 100 instances in each class (i.e., hate, abusive, and normal). This guarantees a more robust evaluation of the different models.

\begin{table*}{}
\centering
\resizebox{\textwidth}{!}{
\begin{tabular}{lrrrrrrrrrrrrrrrr}
\toprule
\textbf{ Model} & \texttt{\textbf{amh}} &\texttt{\textbf{ary}} &\texttt{\textbf{arq}} & \texttt{\textbf{hau}} & \texttt{\textbf{ibo}} & \texttt{\textbf{kin}} & \texttt{\textbf{oro}} & \texttt{\textbf{pcm}} & \texttt{\textbf{som}} & \texttt{\textbf{swa}} & \texttt{\textbf{tir}} & \texttt{\textbf{twi}} & \texttt{\textbf{xho}} & \texttt{\textbf{yor}} & \texttt{\textbf{zul}} & \textbf{avg.}
  
  \\ \midrule
\multicolumn{17}{c}{\textbf{Monolingual}}\\
\midrule
\textbf{AfriBERTa} & 69.54 & 67.93 & 30.48 & 82.28 & 89.53 & 79.43 & 73.43 & 66.90 & 65.52 & 91.36 & 73.07 & 74.54 & 81.07 & 72.37 & 83.75 & 72.33 \\
\textbf{AfriTeVa V2} & 73.91 & 76.71 & 25.25 & 79.06 & 83.95 & 77.60 & 71.61 & 68.69 & 69.65 & 90.68 & 72.36 & 64.96 & 54.67 & \textbf{79.88} & 69.05 & 68.73 \\
\textbf{AfroXLMR} & 70.65 & 80.16 & 61.18 & 81.93 & 89.30 & \textbf{80.72} & 72.11 & 67.98 & 66.84 & 91.44 & 74.52 & 77.17 & 82.49 & 72.15 & 83.44 & 76.15 \\
\textbf{AfroXLMR-76L} & 74.36 & 80.05 & 53.52 & \textbf{82.78} & 89.59 & 79.58 & 76.63 & 68.38 & 71.09 & \textbf{91.72} & 76.27 & 76.65 & 84.40 & 72.35 & 84.65 & 76.45 \\ \midrule
\multicolumn{17}{c}{\textbf{Multilingual}}\\
\midrule
\textbf{AfroXLMR-76L} & \textbf{75.25} & \textbf{80.76} & \textbf{63.31} & 82.20 & \textbf{89.85} & 79.56 & \textbf{77.62} & \textbf{69.20} & \textbf{72.26} & 91.22 & \textbf{77.55} & \textbf{78.68} & \textbf{86.83} & 74.32 & \textbf{86.81} & \textbf{78.16} \\

\bottomrule
\end{tabular}
}
\caption{Model performances after fine-tuning BERT-based LMs. The best performance for each language is highlighted in \textbf{bold}.}
\label{tab:model_finetune_performance}
\end{table*}

\section{Experiments} 
\subsection{Setup}
We compare the performance of three main sets of approaches on the AfriHate datasets: 
\begin{enumerate}[noitemsep,nolistsep]
    \item Fine-tuning BERT-based pre-trained language model (PLM),
    \item Few-shot learning with SetFit with BERT-based PLM \citep{Tunstall2022EfficientFL}: a few-shot approach using BERT-like PLMs,
    \item Prompting large language models (LLMs) in zero and few-shot settings. 
\end{enumerate}
\subsubsection{Fine-tuning PLMs} We use four widely adopted Africa-centric PLMs that have been shown to consistently perform better than massively multilingual PLMs such as XLM-R~\citep{conneau-etal-2020-unsupervised}. The models are AfriBERTa-large~\citep{ogueji-etal-2021-small}, AfriTeVa V2 base~\citep{oladipo-etal-2023-better}, AfroXLMR~\citep{alabi-etal-2022-adapting} and AfroXLMR-76L~\citep{adelani-etal-2024-sib}. Each model was trained for 20 epochs over 5 runs with a batch size of 32, maximum sequence length of 128, and a learning rate of $5e-5$ except for AfroXLMR-76L where we used a learning rate of $3e-5$. The rest of the hyperparameters were the default values set in the HuggingFace fine-tuning pipeline for text classification.

\subsubsection{SetFit Few-shot Learning} SetFit is a few-shot learning approach based on sentence-transformer models like LabSE~\cite{labse}. It works by, first, fine-tuning a pre-trained sentence transformer model on a few examples in a contrastive manner. Then, the resulting model is used to generate rich text embeddings, which are used to train a classification head. We used LaBSE to train classifiers with the following configurations: 
\begin{enumerate}[noitemsep,nolistsep]
    \item for \textbf{zero-shot learning}, we trained the transformers for one epoch using the dummy dataset generated by the framework (2 x [This sentence is \{Label\}, ...]), where \{Label\} can be ``Neutral'', ``Abusive'' or ``Hate'';
    \item for \textbf{few-shot learning}, we trained each model for three epochs using 5, 10, and 20 shots. All the classifiers were trained using a batch size of 32.
\end{enumerate}

\subsubsection{Prompting LLMs} We prompt one closed model (GPT-4o) and nine open models of various model sizes (0.4B to 70B). The open models are: \inkuba~\citep{Tonja2024InkubaLMAS}, \mto~\citep{muennighoff-etal-2023-crosslingual}, \bloom~\citep{Scao2022BLOOMA1}, \mistral~\citep{Jiang2023Mistral7}, \aya~\citep{Aryabumi2024Aya2O}, LLaMa 3.1 \{8B \& 70B\}~\citep{Dubey2024TheL3}, and Gemma 2 \{9B \& 27B\}~\citep{Riviere2024Gemma2I}. All the models were prompted using five prompt templates with clear definitions of the \textit{Abusive}, \textit{Hate}, and \textit{Neutral} categories. For hate speech, we used the definition adopted by the United Nations and another one from the Merriam Webster's dictionary. We report the average scores across all the templates in the results Section. The full prompts can be found in \Cref{app:prompt_templates}.

\begin{table*}[!htb]
    \centering
    \resizebox{\textwidth}{!}{
    \begin{tabular}{lcrrrlcrrrlcrrr}
    \toprule
    \multirow{2}{*}{\textbf{Lang.}} & \multicolumn{4}{c}{\textbf{Monolingual AfroXLMR-76L}} & & \multicolumn{4}{c}{\textbf{Multilingual AfroXLMR-76L}} & & \multicolumn{4}{c}{\textbf{GPT-4o}} \\ \cmidrule{2-5}\cmidrule{7-10}\cmidrule{12-15}
        & \textbf{Macro F1} & \textbf{Abuse} & \textbf{Hate} & \textbf{Neutral} & & \textbf{Macro F1} & \textbf{Abuse} & \textbf{Hate} & \textbf{Neutral} & & \textbf{Macro F1} & \textbf{Abuse} & \textbf{Hate} & \textbf{Neutral} \\
    \midrule
        \texttt{amh} & 73.83 & 70.20 & 77.27 & 74.02 &  & \textbf{75.55} & \textbf{71.92} & \textbf{78.18} & \textbf{76.54} &  & 65.70 & 59.21 & 67.91 & 70.00 \\
        \texttt{ary} & 78.01 & \textbf{86.81} & 66.98 & 80.25 &  & \textbf{79.05} & 86.59 & \textbf{68.66} & 81.91 &  & 75.93 & 84.29 & 61.09 & \textbf{82.43} \\
        \texttt{arq} & 57.05 & 77.35 & 27.42 & 66.38 &  & 68.99 & \textbf{78.29} & 55.95 & \textbf{72.73} &  & \textbf{73.67} & 77.63 & \textbf{71.43} & 71.96 \\
        \texttt{hau} & \textbf{81.53} & \textbf{77.60} & \textbf{80.45} & \textbf{86.54} &  & 78.05 & 76.45 & 72.83 & 84.88 &  & 59.44 & 70.61 & 40.77 & 66.95 \\
        \texttt{ibo} & 88.00 & 92.18 & \textbf{91.18} & 80.64 &  & \textbf{88.33} & \textbf{92.41} & \textbf{91.18} & \textbf{81.40} &  & 76.85 & 84.13 & 75.37 & 71.04 \\
        \texttt{kin} & \textbf{77.96} & \textbf{70.09} & 80.87 & \textbf{82.90} &  & 77.11 & 70.00 & 78.90 & 82.43 &  & 74.27 & 65.45 & \textbf{82.01} & 75.37 \\
        \texttt{orm} & \textbf{70.07} & 47.19 & \textbf{81.17} & \textbf{81.86} &  & 69.93 & \textbf{48.89} & 80.11 & 80.78 &  & 65.30 & 44.12 & 72.78 & 79.00 \\
        \texttt{pcm} & \textbf{65.40} & \textbf{71.09} & 52.82 & 72.31 &  & 63.71 & 69.66 & 49.10 & 72.39 &  & 63.53 & 57.38 & \textbf{56.67} & \textbf{76.54} \\
        \texttt{som} & 59.53 & 67.00 & 30.77 & 80.83 &  & 60.91 & \textbf{68.81} & 32.94 & \textbf{81.00} &  & \textbf{62.94} & 63.85 & \textbf{44.91} & 80.05 \\
        \texttt{swa} & 88.00 & \textbf{92.18} & \textbf{91.18} & 80.64 & & \textbf{89.50} & 89.35 & 88.33 & \textbf{90.83} &  & 83.29 & 85.48 & 80.49 & 83.91 \\
        \texttt{tir} & 72.32 & 72.29 & 82.60 & 62.07 &  & \textbf{74.45} & \textbf{75.23} & \textbf{84.97} & \textbf{63.16} &  & 56.23 & 51.73 & 64.38 & 52.59 \\
        \texttt{twi} & 58.30 & \textbf{91.45} & \textbf{61.40} & 22.05 &  & \textbf{63.66} & 91.04 & 60.18 & 39.76 &  & 62.64 & 85.90 & 54.45 & \textbf{47.58} \\
        \texttt{xho} & 79.37 & \textbf{86.33} & 66.23 & 85.53 &  & \textbf{81.57} & 85.14 & \textbf{71.95} & \textbf{87.63} &  & 57.74 & 70.82 & 38.46 & 63.94 \\
        \texttt{yor} & 57.36 & \textbf{84.24} & 6.90 & 80.95 &  & 59.41 & 84.03 & 11.76 & \textbf{82.44} &  & \textbf{71.48} & 81.56 & \textbf{53.01} & 79.88 \\
        \texttt{zul} & 80.97 & \textbf{87.54} & 68.79 & 86.57 &  & \textbf{84.34} & 85.46 & \textbf{79.55} & \textbf{88.01} &  & 70.63 & 74.60 & 67.78 & 69.51 \\
    \bottomrule
    \end{tabular}
    }
    \caption{Macro F1 scores per class for monolingual and multilingual AfroXLMR-76L [only one run] vs. GPT-4o [prompt template 1; 20 shots]. The best performance for each language is highlighted in \textbf{bold}.}
    \label{tab:per_class_f1}
\end{table*}

\subsection{Experimental Results}

\begin{table*}[!htb]
\centering
\resizebox{\textwidth}{!}{
\begin{tabular}{llrrrrrrrrrrrrrrr||r}
\toprule
 \textbf{Model} & \textbf{\# Shots} & \texttt{\textbf{amh}} &\texttt{\textbf{ary}} &\texttt{\textbf{arq}} & \texttt{\textbf{hau}} & \texttt{\textbf{ibo}} & \texttt{\textbf{kin}} & \texttt{\textbf{oro}} & \texttt{\textbf{pcm}} & \texttt{\textbf{som}} & \texttt{\textbf{swa}} & \texttt{\textbf{tir}} & \texttt{\textbf{twi}} & \texttt{\textbf{xho}} & \texttt{\textbf{yor}} & \texttt{\textbf{zul}} & \textbf{Avg.}
  
  \\ \midrule
\textbf{SetFit} & 0 & 33.79 & 46.54 & 27.4 & 23.57 & 36.90 & 25.29 & 26.26 & 43.46 & 32.18 & 44.38 & 49.30 & 51.13 & 37.38 & 48.46 & 35.08 & 37.41 \\
 & 5 & 44.89 & 33.82 & 36.07 & 49.93 & 51.52 & 55.35 & 36.56 & 53.28 & 48.09 & 54.90 & 35.36 & 33.23 & 47.49 & 34.36 & 33.18 & 43.20 \\
 & 10 & 49.42 & 35.81 & 46.55 & 49.72 & 39.31 & 53.10 & 39.10 & 57.00 & 42.29 & 62.64 & 39.16 & 45.76 & 43.52 & 49.73 & 40.33 & 46.23 \\
 & 20 & 50.13 & 40.49 & 54.24 & 55.40 & 65.15 & 57.35 & 40.91 & 59.09 & 48.95 & 74.93 & 40.04 & 53.72 & 46.92 & 56.54 & 50.67 & 52.97 \\\midrule
\textbf{Mistral-7B-v0.1} & 0 & 21.18 & 17.46 & 11.98 & 6.84 & 8.04 & 15.18 & 20.82 & 6.76 & 8.16 & 18.88 & 25.62 & 8.38 & 9.72 & 8.22 & 8.92 & 13.08 \\
 & 5 & 37.68 & 31.60 & 44.54 & 34.90 & 36.84 & 46.98 & 39.46 & 50.00 & 34.12 & 58.58 & 35.64 & 32.04 & 34.24 & 41.04 & 32.54 & 39.35 \\
 & 10 & 39.68 & 36.50 & 50.56 & 37.82 & 36.94 & 45.72 & 39.02 & 52.10 & 31.82 & 63.18 & 35.84 & 30.82 & 32.64 & 42.90 & 32.46 & 40.53 \\
 & 20 & 36.64 & 35.46 & 52.94 & 38.44 & 38.98 & 52.28 & 39.24 & 54.60 & 33.28 & 66.28 & 37.42 & 30.44 & 34.12 & 46.40 & 35.00 & 42.10 \\\midrule
\textbf{aya-23-35B} & 0 & 17.04 & 19.20 & 17.20 & 20.34 & 14.04 & 21.34 & 20.38 & 20.02 & 18.56 & 34.04 & 17.10 & 10.68 & 20.32 & 17.48 & 20.24 & 19.20 \\
 & 5 & 37.78 & 56.40 & 57.00 & 39.70 & 42.70 & 48.78 & 39.24 & 57.90 & 38.72 & 69.84 & 35.68 & 40.24 & 36.76 & 48.88 & 44.58 & 46.28 \\
 & 10 & 38.50 & 53.96 & 61.82 & 44.42 & 45.80 & 52.82 & 42.16 & 59.26 & 37.60 & 74.10 & 35.88 & 39.92 & 36.18 & 48.44 & 45.00 & 47.72 \\
 & 20 & 38.10 & 52.90 & 63.76 & 46.14 & 46.52 & 57.62 & 42.26 & 61.68 & 38.68 & 75.68 & 38.12 & 38.54 & 37.74 & 51.52 & 44.52 & 48.92 \\\midrule
\textbf{Gemma-2-9B} & 0 & 27.42 & 28.36 & 25.46 & 14.48 & 15.00 & 24.08 & 23.74 & 24.74 & 10.92 & 37.12 & 25.52 & 19.50 & 13.40 & 21.66 & 12.52 & 21.59 \\
 & 5 & 56.60 & 57.68 & 61.14 & 49.98 & 48.12 & 60.26 & 45.10 & 60.30 & 46.42 & 74.78 & 48.40 & 46.50 & 35.82 & 55.96 & 46.60 & 52.91 \\
 & 10 & 57.84 & 61.08 & 61.72 & 56.62 & 53.78 & 63.54 & 45.78 & 60.54 & 51.72 & 78.16 & 45.48 & 48.16 & 38.82 & 60.22 & 52.58 & 55.74 \\
 & 20 & 60.96 & 59.94 & 64.38 & 55.90 & 54.56 & 64.14 & 44.70 & 62.54 & 52.22 & 79.24 & 44.94 & 48.92 & 39.02 & 60.50 & 53.68 & 56.38 \\\midrule
\textbf{Gemma-2-27B} & 0 & 41.78 & 41.24 & 41.12 & 28.96 & 31.36 & 42.08 & 33.46 & 49.78 & 31.10 & 59.88 & 30.84 & 27.74 & 25.64 & 41.46 & 28.02 & 36.96 \\
 & 5 & 59.62 & 64.14 & 65.62 & 54.62 & 56.14 & 61.08 & 46.76 & 61.78 & 54.12 & 81.18 & 52.12 & 47.26 & 40.48 & 61.72 & 54.28 & 57.39 \\
 & 10 & 60.70 & 61.36 & 64.88 & 58.90 & 56.84 & 64.86 & 46.96 & 61.60 & 52.86 & 81.94 & 49.82 & 50.86 & 41.90 & 59.66 & 56.94 & 58.01 \\
 & 20 & 62.28 & 59.86 & 65.80 & 59.60 & 58.76 & 66.06 & 48.90 & 63.24 & 52.90 & 82.98 & 53.36 & 50.34 & 43.08 & 59.10 & 56.88 & 58.88 \\\midrule
\textbf{Llama-3.1-70B} & 0 & 36.34 & 43.34 & 42.64 & 35.64 & 32.52 & 38.52 & 31.54 & 48.66 & 31.14 & 60.14 & 27.08 & 25.54 & 28.98 & 40.64 & 35.50 & 37.21 \\
 & 5 & 58.52 & 66.24 & 62.88 & 52.86 & 52.88 & 58.14 & 45.50 & 64.38 & 52.72 & 75.76 & 44.42 & 43.90 & 39.02 & 58.08 & 55.24 & 55.37 \\
 & 10 & 61.18 & 64.46 & 62.20 & 56.40 & 55.10 & 59.00 & 46.60 & 63.58 & 54.40 & 78.74 & 49.62 & 45.76 & 39.04 & 57.96 & 55.74 & 56.65 \\
 & 20 & 60.38 & 61.36 & 63.80 & 57.40 & 53.80 & 62.48 & 49.02 & 63.18 & 51.82 & 80.02 & 53.90 & 47.50 & 40.34 & 57.94 & 56.36 & 57.29 \\\midrule
\textbf{GPT-4o} & 0 & 61.78 & 66.41 & 73.75 & 56.91 & 68.82 & 62.53 & 60.01 & 65.94 & 63.42 & 73.66 & 45.75 & 52.72 & 58.92 & 75.21 & 54.47 & 62.69 \\
 & 5 & 66.73 & 73.53 & 77.33 & 55.44 & 76.73 & 72.27 & 70.94 & 66.29 & 59.33 & 80.95 & 61.11 & 73.21 & 60.45 & 76.98 & 59.34 & 68.71 \\
 & 10 & 67.94 & 75.54 & 77.54 & 58.15 & 80.08 & 75.07 & 72.27 & 67.14 & 62.75 & 84.19 & 57.52 & 72.28 & 65.75 & 76.74 & 65.05 & 70.53 \\
 & 20 & 68.08 & 76.16 & 78.69 & 58.34 & 80.81 & 74.86 & 72.33 & 65.41 & 66.44 & 84.61 & 59.55 & 75.86 & 66.08 & 77.11 & 71.36 & 71.71 \\\midrule\midrule
\textbf{Lang. avg.} & - & 48.32 & 50.74 & 54.04 & 44.91 & 47.79 & 52.88 & 43.18 & 54.44 & 43.10 & 67.53 & 41.95 & 42.53 & 39.06 & 51.25 & 44.18 & 48.39 \\

\bottomrule[1.5pt] 
\end{tabular}
}
\caption{\textbf{Model performance (Macro F1-score) for zero- and-few shot classifiers} across the different languages in AfriHate.
The best performance for each language is highlighted in \textbf{bold}. This is an average over 5 prompt templates.}
\label{tab:llm_performance}
\end{table*}
\subsubsection{Monolingual vs. Multilingual Fine-tuning}
We compare monolingual fine-tuning where we train on a language and evaluate on the same language to multilingual fine-tuning---where we combine the training data of all the languages and evaluate the results for each.

\autoref{tab:model_finetune_performance} shows the results of the fine-tuning experiments. We find that encoder-only models perform better than the T5-style models, i.e., AfriTeVa V2. On average, AfroXLMR-76L achieved the best performance, most likely due to the fact that it was pre-trained on all the languages included in AfriHate. While AfroXLMR was not pre-trained on some languages such as Tigrinya (\texttt{tir}) and Twi (\texttt{twi}), it still achieves performance that is comparable to AfroXLMR-76L. 
AfriBERTa generally struggles with Arabic dialects such as Algerian Darja (\texttt{arq}) and Moroccan Darija (\texttt{ary}) as the Arabic script was not included in its pre-training. 

Overall, multilingual training of a single model leads to the best results on 11 out of 14 languages, and comparable results on the remaining languages except for \yoruba(\texttt{yor}), where AfriBERTa led to the best result because of its Africa-centric tokenizer. 

\autoref{tab:per_class_f1} shows the per-class accuracy across different languages. Overall, multilingual models perform better for languages with a low percentage of the ``hate'' category in the training data (e.g., $<200$) such as \texttt{ary}, \texttt{xho}, \texttt{yor}, and \texttt{zul} with an F-score improvement of $+1.7$, $+5.7$, $+4.9$, and $+10.8$, respectively. 

\subsubsection{Zero-shot vs. Few-shot Settings}
\autoref{tab:llm_performance} shows the results of both zero-shot and few-shot experiments. SetFit performs slightly better than all open LLMs in zero-shot settings ($36.9$), and GPT-4o leads to the best overall performance with $61.9$ F1 points. 

When considering the few-shot settings, 5-shot models show the biggest boost in performance, where \gemma and other bigger models (\gemmaTwoL and \llamal) improve by about $+20$ points. The performance boost with additional 10 and 20 shots is more limited for LLMs ($+3.0$ improvement), whereas SetFit consistently benefits from additional examples. 
The best results reached for closed models are at 20-shots, where \gpto achieved an overall F1-score of $70.8$ while \gemmaTwoL achieved the best overall results for any open model with an F1-score of $57.2$.

In our performance analysis per different classes shown in \autoref{tab:per_class_f1}, \gpto generally performs worse than full fine-tuning in monolingual or multilingual settings. However, we find that it achieves significantly better performance for hate detection in a few languages such as \texttt{arq} ($+44.0$), \texttt{som} ($+14.9$), and \texttt{yor} ($+46.1$) compared to monolingual fine-tuning. For languages without enough training data for the hate category such as \yoruba, prompting LLMs might provide a better detection of this class compared to fine-tuning BERT-like PLMs. 

\subsubsection{Overall Results}
Overall, our results show that fine-tuning multilingual models leads to a better performance for the majority of the AfriHate languages. That is, AfroXLMR-76L achieves an average macro F1 score of \textbf{78.16}. 
As for zero-shot and few-shot settings,~\textbf{GPT-4o} outperformed other models, with average F1 scores of \textbf{61.89} and \textbf{70.79} in zero-shot and 20-shot settings, respectively. 

Overall, these results highlight the advantages of multilingual and context-specific models in hate and abusive language detection for African languages.

\section{Conclusion} 
\label{sec:Conclusion}

We introduced AfriHate, the first large-scale collection of hate and abusive language datasets in 15 African languages: Algerian Arabic, Amharic, Igbo, Kinyarwanda, Hausa, Moroccan Arabic, Nigerian Pidgin, Oromo, Somali, Swahili, Tigrinya, Twi, Xhoza, \yoruba, and Zulu. The datasets were annotated by native speakers as hate speech, abusive, or neutral. 
We discussed our data collection strategies and highlighted the challenges faced during the data collection and annotation.
We then reported baseline experiments using Africa-centric pre-trained language models as well as prompted open and closed LLMs showing a large gap in the performance across languages. 

AfriHate is a first step towards building high-quality hate speech resources for African languages. We publicly release all the datasets, scripts, models, and lexicons to the research community.

\section{Limitations}
While we collected AfriHate using large sets of keywords, we acknowledge the unavoidable presence of selection bias \cite{ousidhoum2020comparative} as no dataset can capture the full range of hate speech contexts across various languages and cultures.
In addition, although we recruited annotators who come from different socio-cultural backgrounds, hate speech remains a subjective task and one cannot include all possible perspectives of what constitutes hate speech or abuse. We mitigate the problem by sharing the individual annotations with the research community studying the problem.

Further, when using language identification to collect data, challenges due to code-mixing and digraphia, make the task non-trivial given the common usage of multilinguality in African languages. We address the problem by asking the annotators to flag any tweet that is not in the target language. We acknowledge, nevertheless, instances that may have been missed by our annotators.

Finally, we report on the various dataset statistics and model features. However, given the fact that we use some closed models in our experiments, and the class imbalance problem which is inherent to hate speech datasets, we do not claim that our results are fully replicable or generalisable. 

\section{Ethical Considerations} 

\paragraph{Annotators} The annotators involved in this study were compensated for their work by more than the minimum wage and any demographic information about them was shared with consent.
We acknowledge the difficulty of annotating hate speech and abusive language on people's well-being. Therefore, the annotators could reach out to us and were allowed to quit at any time.

\paragraph{Language Use}
Our datasets focus on hate speech and abusive languages in 15 African languages. However, we do not claim that our datasets represent the full usage of these languages. We further acknowledge the socio-cultural biases that can come with the data as views on hate highly differ from one person to another and those shared by our annotators cannot include all possible perspectives. 

\paragraph{Intended uses and potential misuses} 
Our datasets focus on hate speech and abusive language. They present a first step towards studying the phenomenon in some low-resource African languages. However, as malicious data actors can misuse our resources, we follow the suggestions made by \citet{schlichtkrull2023intended} for automated fact-checking researchers and clearly state the following:
\begin{itemize}[noitemsep,nolistsep]
    \item Models built using our datasets \textbf{should not be used for automated removal}.
    \item Our \textit{data subjects} are social media users.
    \item Our \textit{data actors} and \textit{model owners} should be users, moderators, experts, and researchers with background knowledge in the field, especially on the limitations of automated hate speech and abusive language detection models.
    \item Given the sensitivity of the task and the high risk of false positives, any constructed or deployed model using our data should be \textbf{human-in-the-loop} with the humans being native or near-native speakers.
\end{itemize}

\section*{Acknowledgments}
We are grateful for the work and insights shared by our annotators.

This work was carried out with support from Lacuna Fund, an initiative co-founded by The Rockefeller Foundation, Google.org, and Canada’s International Development Research Centre. The views expressed herein do not necessarily represent those of Lacuna Fund, its Steering Committee, its funders, or Meridian Institute.  We thank Label Studio for providing the academic version annotation tool used in this project.
We also thank OpenAI for providing API credits to Masakhane. Shamsuddeen acknowledges the support of Google DeepMind.
Paul was supported by a MUR FARE 2020 initiative under grant agreement Prot.\ R20YSMBZ8S (INDOMITA) and the European Research Council (ERC) under the European Union’s Horizon 2020 research and innovation program (No.\ 949944, INTEGRATOR).
Abinew acknowledges the support of LT group, University of Hamburg.

\bibliography{custom}

\clearpage
\appendix

\section{Example Appendix}
\label{sec:appendix}

\subsection{Prompt templates}
\label{app:prompt_templates}
\begin{tcolorbox}[colframe=blue!50!black, title=Prompt 1]
I am providing you with the definition of Hate speech, Abusive language, and Neutral tweets.

\vspace{0.3cm}

Hate speech is language content that expresses hatred towards a particular group or individual based on their political affiliation, race, ethnicity, religion, gender, sexual orientation, or other characteristics. It also includes threats of violence.

\vspace{0.3cm}

Neutral language does not contain any bad language.

\vspace{0.3cm}
Which category does the tweet above belong to: 'Hate', 'Abuse', or 'Neutral'? Pick exactly one category. Don't give any additional context, just classify the tweet.


\vspace{0.3cm}
Tweet: {text}
Category:

\end{tcolorbox}


\begin{tcolorbox}[colframe=blue!50!black, title=Prompt 2]
Read the following label definitions and provide a label without any explanations.

\vspace{0.3cm}

Hate: Hate speech is public speech that expresses hate or encourages violence towards a person or group based on something such as race, religion, gender, ethnicity, sexual orientation or other characteristics.

\vspace{0.3cm}

Abusive: Abusive and offensive language means verbal messages that use words in an inappropriate way and may include but is not limited to swearing, name-calling, or profanity. Offensive language may upset or embarrass people because it is rude or insulting

\vspace{0.3cm}
Neutral: Neutral language is neither hateful nor abusive or offensive. It does not contain any bad language.


\vspace{0.3cm}

Text: {tweet}

Label: 
\end{tcolorbox}

\begin{tcolorbox}[colframe=blue!50!black, title=Prompt 3]
Read the following text and definitions:

\vspace{0.3cm}

Text: {tweet}.

\vspace{0.3cm}

Definitions:
Hate: Hate speech is public speech that expresses hate or encourages violence towards a person or group based on something such as race, religion, gender, ethnicity, sexual orientation or other characteristics.

\vspace{0.3cm}
Abuse: Abusive and offensive language means verbal messages that use words in an inappropriate way and may include but is not limited to swearing, name-calling, or profanity. Offensive language may upset or embarrass people because it is rude or insulting


\vspace{0.3cm}

Neutral: Neutral language is neither hateful nor abusive or offensive. It does not contain any bad language.
\vspace{0.3cm}
Which of these definitions (hate, abuse, neutral) apply to this tweet?

\end{tcolorbox}

\begin{tcolorbox}[colframe=blue!50!black, title=Prompt 4]
Read the following definitions and text to categorize:

\vspace{0.3cm}

Definitions:
Hate: Hate speech is public speech that expresses hate or encourages violence towards a person or group based on something such as race, religion, gender, ethnicity, sexual orientation or other characteristics.

\vspace{0.3cm}

Abuse: Abusive and offensive language means verbal messages that use words in an inappropriate way and may include but is not limited to swearing, name-calling, or profanity. Offensive language may upset or embarrass people because it is rude or insulting

\vspace{0.3cm}
Neutral: Neutral language is neither hateful nor abusive or offensive. It does not contain any bad language.
  

\vspace{0.3cm}
Text: {tweet}.
\vspace{0.3cm}
Which of these definitions (hate, abuse, neutral) apply to this tweet?

\end{tcolorbox}

\begin{tcolorbox}[colframe=blue!50!black, title=Prompt 5]
You will be given a text snippet and 3 category definitions. 
Your task is to choose which category applies to this text. 

\vspace{0.3cm}

Your text snippet is: {tweet}

\vspace{0.3cm}

Your category definitions are:

\vspace{0.3cm}
HATE category definition: Hate speech is public speech that expresses hate or encourages violence towards a person or group based on something such as race, religion, gender, ethnicity, sexual orientation or other characteristics.
 
\vspace{0.3cm}
ABUSE category definition: Abusive and offensive language means verbal messages that use words in an inappropriate way and may include but is not limited to swearing, name-calling, or profanity. Offensive language may upset or embarrass people because it is rude or insulting
 
\vspace{0.3cm}
NEUTRAL category definition: Neutral language is neither hateful nor abusive or offensive. It does not contain any bad language.
 
\vspace{0.3cm}

Does the text snippet belong to the HATE, ABUSIVE, or the NEUTRAL category?
Thinking step by step answer HATE, ABUSIVE, or NEUTRAL capitalizing all the letters. 
Explain your reasoning FIRST, then output HATE, ABUSIVE, or NEUTRAL.
\end{tcolorbox}



\begin{table*}
\small
\centering
\resizebox{\textwidth}{!}{
\begin{tabular}{llrrrrrrrrrrrrrrr||r}
\toprule
 \textbf{Model} & \textbf{\# Shots} & \texttt{\textbf{amh}} &\texttt{\textbf{ary}} &\texttt{\textbf{arq}} & \texttt{\textbf{hau}} & \texttt{\textbf{ibo}} & \texttt{\textbf{kin}} & \texttt{\textbf{oro}} & \texttt{\textbf{pcm}} & \texttt{\textbf{som}} & \texttt{\textbf{swa}} & \texttt{\textbf{tir}} & \texttt{\textbf{twi}} & \texttt{\textbf{xho}} & \texttt{\textbf{yor}} & \texttt{\textbf{zul}} & \textbf{Avg.}
  
  \\ \midrule
\textbf{SetFit} & 0 & 33.79 & 46.54 & 27.40 & 23.57 & 36.90 & 25.29 & 26.26 & 43.46 & 32.18 & 44.38 & 49.30 & 51.13 & 37.38 & 48.46 & 35.08 & 37.41 \\
   & 5 & 44.89 & 33.82 & 36.07 & 49.93 & 51.52 & 55.35 & 36.56 & 53.28 & 48.09 & 54.90 & 35.36 & 33.23 & 47.49 & 34.36 & 33.18 & 43.20 \\
   & 10 & 49.42 & 35.81 & 46.55 & 49.72 & 39.31 & 53.10 & 39.10 & 57.00 & 42.29 & 62.64 & 39.16 & 45.76 & 43.52 & 49.73 & 40.33 & 46.23 \\
   & 20 & 50.13 & 40.49 & 54.24 & 55.40 & 65.15 & 57.35 & 40.91 & 59.09 & 48.95 & 74.93 & 40.04 & 53.72 & 46.92 & 56.54 & 50.67 & 52.97 \\\midrule
  \textbf{InkubaLM-0.4B} & 0 & 22.96 & 19.38 & 18.14 & 20.02 & 18.42 & 22.44 & 26.16 & 16.86 & 20.76 & 20.74 & 25.10 & 17.44 & 20.58 & 19.26 & 20.38 & 20.58 \\
   & 5 & 28.04 & 24.70 & 20.06 & 21.78 & 21.98 & 25.78 & 26.86 & 19.08 & 25.06 & 20.48 & 29.26 & 24.50 & 23.04 & 20.98 & 21.68 & 23.55 \\
   & 10 & 27.00 & 26.46 & 23.92 & 24.48 & 23.52 & 27.22 & 29.14 & 23.68 & 25.98 & 24.58 & 27.92 & 24.02 & 24.68 & 23.68 & 25.06 & 25.42 \\
   & 20 & 27.06 & 26.42 & 24.60 & 24.30 & 23.86 & 26.96 & 28.58 & 23.76 & 25.88 & 24.64 & 30.04 & 23.32 & 23.56 & 23.14 & 25.20 & 25.42 \\\midrule
  \textbf{mt0-small} & 0 & 18.86 & 16.98 & 13.42 & 12.22 & 9.72 & 17.40 & 20.68 & 12.40 & 15.16 & 14.07 & 19.30 & 8.56 & 14.20 & 12.04 & 13.62 & 14.58 \\
   & 5 & 26.38 & 25.76 & 21.66 & 19.94 & 21.28 & 23.04 & 27.78 & 25.98 & 21.40 & 20.30 & 26.20 & 12.66 & 19.96 & 17.72 & 19.50 & 21.97 \\
   & 10 & 27.38 & 23.72 & 20.40 & 20.94 & 22.36 & 21.08 & 27.94 & 25.58 & 19.88 & 16.93 & 26.62 & 12.08 & 18.56 & 19.90 & 18.88 & 21.48 \\
   & 20 & 26.28 & 23.38 & 19.86 & 21.60 & 22.14 & 21.18 & 28.16 & 25.06 & 19.96 & 20.05 & 28.04 & 13.44 & 19.12 & 19.76 & 18.50 & 21.77 \\\midrule
  \textbf{bloomz-7b1-mt} & 0 & 17.18 & 21.86 & 21.86 & 18.60 & 18.42 & 21.98 & 19.38 & 21.60 & 16.74 & 25.10 & 16.92 & 17.70 & 19.24 & 22.36 & 19.80 & 19.92 \\
   & 5 & 22.74 & 21.40 & 25.88 & 23.10 & 27.88 & 25.36 & 26.22 & 26.66 & 23.54 & 27.03 & 25.64 & 28.34 & 23.26 & 24.80 & 22.62 & 24.96 \\
   & 10 & 27.86 & 21.06 & 25.88 & 24.40 & 27.90 & 22.54 & 27.86 & 26.24 & 25.48 & 19.83 & 28.58 & 28.50 & 23.56 & 24.16 & 23.92 & 25.18 \\
   & 20 & 25.46 & 19.92 & 24.64 & 24.96 & 27.72 & 24.18 & 26.64 & 26.82 & 25.76 & 24.50 & 28.48 & 28.84 & 23.24 & 25.28 & 23.88 & 25.35 \\\midrule
  \textbf{Mistral-7B-v0.1} & 0 & 21.18 & 17.46 & 11.98 & 6.84 & 8.04 & 15.18 & 20.82 & 6.76 & 8.16 & 10.64 & 25.62 & 8.38 & 9.72 & 8.22 & 8.92 & 12.53 \\
   & 5 & 37.68 & 31.60 & 44.54 & 34.90 & 36.84 & 46.98 & 39.46 & 50.00 & 34.12 & 49.58 & 35.64 & 32.04 & 34.24 & 41.04 & 32.54 & 38.75 \\
   & 10 & 39.68 & 36.50 & 50.56 & 37.82 & 36.94 & 45.72 & 39.02 & 52.10 & 31.82 & 54.34 & 35.84 & 30.82 & 32.64 & 42.90 & 32.46 & 39.94 \\
   & 20 & 36.64 & 35.46 & 52.94 & 38.44 & 38.98 & 52.28 & 39.24 & 54.60 & 33.28 & 55.98 & 37.42 & 30.44 & 34.12 & 46.40 & 35.00 & 41.41 \\\midrule
  \textbf{aya-23-35B} & 0 & 17.04 & 19.20 & 17.20 & 20.34 & 14.04 & 21.34 & 20.38 & 20.02 & 18.56 & 21.20 & 17.10 & 10.68 & 20.32 & 17.48 & 20.24 & 18.34 \\
   & 5 & 37.78 & 56.40 & 57.00 & 39.70 & 42.70 & 48.78 & 39.24 & 57.90 & 38.72 & 56.32 & 35.68 & 40.24 & 36.76 & 48.88 & 44.58 & 45.38 \\
   & 10 & 38.50 & 53.96 & 61.82 & 44.42 & 45.80 & 52.82 & 42.16 & 59.26 & 37.60 & 61.08 & 35.88 & 39.92 & 36.18 & 48.44 & 45.00 & 46.86 \\
   & 20 & 38.10 & 52.90 & 63.76 & 46.14 & 46.52 & 57.62 & 42.26 & 61.68 & 38.68 & 73.68 & 38.12 & 38.54 & 37.74 & 51.52 & 44.52 & 48.79 \\\midrule
  \textbf{Gemma-2-9B} & 0 & 27.42 & 28.36 & 25.46 & 14.48 & 15.00 & 24.08 & 23.74 & 24.74 & 10.92 & 32.12 & 25.52 & 19.50 & 13.40 & 21.66 & 12.52 & 21.26 \\
   & 5 & 56.60 & 57.68 & 61.14 & 49.98 & 48.12 & 60.26 & 45.10 & 60.30 & 46.42 & 72.88 & 48.40 & 46.50 & 35.82 & 55.96 & 46.60 & 52.78 \\
   & 10 & 57.84 & 61.08 & 61.72 & 56.62 & 53.78 & 63.54 & 45.78 & 60.54 & 51.72 & 76.50 & 45.48 & 48.16 & 38.82 & 60.22 & 52.58 & 55.63 \\
   & 20 & 60.96 & 59.94 & 64.38 & 55.90 & 54.56 & 64.14 & 44.70 & 62.54 & 52.22 & 77.56 & 44.94 & 48.92 & 39.02 & 60.50 & 53.68 & 56.26 \\\midrule
  \textbf{Gemma-2-27B} & 0 & 41.78 & 41.24 & 41.12 & 28.96 & 31.36 & 42.08 & 33.46 & 49.78 & 31.10 & 54.54 & 30.84 & 27.74 & 25.64 & 41.46 & 28.02 & 36.61 \\
   & 5 & 59.62 & 64.14 & 65.62 & 54.62 & 56.14 & 61.08 & 46.76 & 61.78 & 54.12 & 77.96 & 52.12 & 47.26 & 40.48 & 61.72 & 54.28 & 57.18 \\
   & 10 & 60.70 & 61.36 & 64.88 & 58.90 & 56.84 & 64.86 & 46.96 & 61.60 & 52.86 & 80.04 & 49.82 & 50.86 & 41.90 & 59.66 & 56.94 & 57.88 \\
   & 20 & 62.28 & 59.86 & 65.80 & 59.60 & 58.76 & 66.06 & 48.90 & 63.24 & 52.90 & 81.18 & 53.36 & 50.34 & 43.08 & 59.10 & 56.88 & 58.76 \\\midrule
  \textbf{Llama-3.1-8B} & 0 & 17.36 & 19.34 & 18.82 & 24.36 & 12.70 & 23.90 & 22.56 & 20.10 & 24.86 & 20.82 & 14.18 & 9.02 & 21.42 & 18.60 & 21.94 & 19.33 \\
   & 5 & 38.08 & 40.88 & 51.68 & 36.78 & 35.90 & 38.78 & 31.24 & 52.38 & 30.62 & 58.10 & 30.20 & 34.82 & 29.22 & 40.42 & 30.90 & 38.67 \\
   & 10 & 42.04 & 43.10 & 54.76 & 37.78 & 37.78 & 43.02 & 35.92 & 53.88 & 32.98 & 64.96 & 31.92 & 34.52 & 28.20 & 38.98 & 30.72 & 40.70 \\
   & 20 & 47.74 & 40.84 & 57.92 & 41.64 & 37.76 & 48.86 & 39.78 & 55.36 & 30.26 & 69.38 & 37.12 & 33.60 & 27.84 & 42.68 & 29.12 & 42.66 \\\midrule
  \textbf{Llama-3.1-70B} & 0 & 36.34 & 43.34 & 42.64 & 35.64 & 32.52 & 38.52 & 31.54 & 48.66 & 31.14 & 49.36 & 27.08 & 25.54 & 28.98 & 40.64 & 35.50 & 36.50 \\
   & 5 & 58.52 & 66.24 & 62.88 & 52.86 & 52.88 & 58.14 & 45.50 & 64.38 & 52.72 & 73.48 & 44.42 & 43.90 & 39.02 & 58.08 & 55.24 & 55.22 \\
   & 10 & 61.18 & 64.46 & 62.20 & 56.40 & 55.10 & 59.00 & 46.60 & 63.58 & 54.40 & 76.50 & 49.62 & 45.76 & 39.04 & 57.96 & 55.74 & 56.50 \\
   & 20 & 60.38 & 61.36 & 63.80 & 57.40 & 53.80 & 62.48 & 49.02 & 63.18 & 51.82 & 77.72 & 53.90 & 47.50 & 40.34 & 57.94 & 56.36 & 57.13 \\\midrule
  \textbf{GPT-4o} & 0 & 61.78 & 66.41 & 73.75 & 56.91 & 68.82 & 62.53 & 60.01 & 65.94 & 63.42 & 73.66 & 45.75 & 52.72 & 58.68 & 75.21 & 54.47 & 62.67 \\
   & 5 & 66.73 & 73.53 & 77.33 & 55.44 & 76.73 & 72.27 & 70.94 & 66.29 & 59.33 & 80.95 & 61.11 & 73.21 & 60.45 & 76.98 & 59.34 & 68.71 \\
   & 10 & 67.94 & 75.54 & 77.54 & 58.15 & 80.08 & 75.07 & 72.27 & 67.14 & 62.75 & 84.19 & 57.52 & 72.28 & 65.75 & 76.74 & 65.05 & 70.53 \\
   & 20 & 68.08 & 76.16 & 78.69 & 58.34 & 80.81 & 74.86 & 72.33 & 65.41 & 66.44 & 84.61 & 59.55 & 75.86 & 66.08 & 77.11 & 71.36 & 71.71 \\\midrule\midrule
  \textbf{Lang. avg.} & - & 40.80 & 41.73 & 44.47 & 37.60 & 39.26 & 43.51 & 37.59 & 44.99 & 36.16 & 51.01 & 36.37 & 35.05 & 33.03 & 41.56 & 36.43 & 39.97 \\

\bottomrule[1.5pt] 
\end{tabular}
}
\caption{\textbf{Model performance (Macro F1-score) for zero- and few shot classifiers} across the 14 languages in AfriHate.
Best performance for each language is highlighted in \textbf{bold}.}
\label{tab:llm_performance_f1}
\end{table*}

\begin{table*}
\centering
\resizebox{\textwidth}{!}{
\begin{tabular}{llrrrrrrrrrrrrrrr||r}
\toprule
 \textbf{model} & \textbf{\#shots} & \texttt{\textbf{amh}} &\texttt{\textbf{ary}} &\texttt{\textbf{arq}} & \texttt{\textbf{hau}} & \texttt{\textbf{ibo}} & \texttt{\textbf{kin}} & \texttt{\textbf{oro}} & \texttt{\textbf{pcm}} & \texttt{\textbf{som}} & \texttt{\textbf{swa}} & \texttt{\textbf{tir}} & \texttt{\textbf{twi}} & \texttt{\textbf{xho}} & \texttt{\textbf{yor}} & \texttt{\textbf{zul}} & \texttt{\textbf{Avg.}} 
  
  \\ \midrule
\textbf{SetFit} & 0 & 33.20 & 42.92 & 26.63 & 24.02 & 38.61 & 21.29 & 28.72 & 39.23 & 29.26 & 35.26 & 47.32 & 52.15 & 37.62 & 41.27 & 35.03 & 35.50 \\
 & 5 & 45.11 & 35.91 & 31.89 & 50.81 & 50.30 & 51.26 & 36.50 & 52.54 & 49.93 & 54.83 & 31.63 & 38.40 & 46.14 & 31.62 & 35.71 & 42.84 \\
 & 10 & 49.26 & 37.05 & 44.27 & 50.43 & 40.80 & 48.60 & 42.56 & 56.87 & 40.27 & 62.88 & 39.74 & 49.28 & 42.60 & 49.21 & 42.03 & 46.39 \\
 & 20 & 50.33 & 40.92 & 53.25 & 54.34 & 65.65 & 57.28 & 43.08 & 58.44 & 48.86 & 74.68 & 38.95 & 55.87 & 46.78 & 53.85 & 51.10 & 52.89 \\\midrule
\textbf{InkubaLM-0.4B} & 0 & 39.74 & 23.62 & 30.26 & 22.66 & 23.04 & 30.96 & 41.14 & 19.84 & 28.30 & 27.02 & 47.24 & 23.16 & 27.30 & 23.70 & 27.38 & 29.02 \\
 & 5 & 39.12 & 23.72 & 31.88 & 25.28 & 34.36 & 32.60 & 44.88 & 23.74 & 30.86 & 25.62 & 48.12 & 43.96 & 28.14 & 26.88 & 27.52 & 32.45 \\
 & 10 & 37.20 & 29.52 & 32.64 & 38.38 & 27.74 & 37.58 & 43.26 & 28.82 & 41.26 & 31.22 & 45.84 & 32.90 & 34.22 & 31.54 & 35.76 & 35.19 \\
 & 20 & 37.64 & 29.18 & 32.68 & 38.64 & 28.70 & 37.84 & 43.06 & 28.98 & 41.34 & 31.37 & 48.88 & 32.30 & 33.82 & 30.84 & 36.12 & 35.43 \\\midrule
\textbf{mt0-small} & 0 & 38.14 & 23.36 & 31.76 & 28.70 & 17.02 & 35.56 & 43.74 & 22.70 & 31.34 & 27.47 & 42.80 & 14.66 & 28.32 & 23.16 & 27.62 & 29.09 \\
 & 5 & 36.70 & 24.90 & 34.12 & 29.18 & 29.48 & 31.36 & 41.10 & 32.06 & 25.26 & 27.90 & 42.56 & 16.20 & 33.04 & 28.38 & 32.86 & 31.01 \\
 & 10 & 37.40 & 25.74 & 35.06 & 26.36 & 33.44 & 28.30 & 40.46 & 31.98 & 24.00 & 23.37 & 43.04 & 15.88 & 31.32 & 29.72 & 32.00 & 30.54 \\
 & 20 & 36.50 & 25.98 & 35.08 & 26.20 & 34.50 & 28.44 & 40.30 & 31.54 & 23.86 & 23.40 & 44.76 & 17.08 & 31.68 & 28.92 & 32.30 & 30.70 \\\midrule
\textbf{bloomz-7b1-mt} & 0 & 34.62 & 34.38 & 33.08 & 32.66 & 29.18 & 36.20 & 37.30 & 32.76 & 33.90 & 38.53 & 35.82 & 28.88 & 31.94 & 33.70 & 32.24 & 33.68 \\
 & 5 & 32.36 & 37.34 & 46.74 & 37.52 & 62.56 & 30.88 & 31.96 & 48.14 & 33.90 & 40.97 & 36.68 & 68.50 & 39.36 & 48.86 & 39.64 & 42.36 \\
 & 10 & 35.16 & 36.96 & 45.92 & 37.72 & 62.40 & 27.74 & 33.06 & 47.10 & 35.04 & 36.97 & 38.70 & 67.94 & 38.56 & 47.66 & 39.44 & 42.02 \\
 & 20 & 33.16 & 35.60 & 44.18 & 37.88 & 63.12 & 29.12 & 31.82 & 47.04 & 34.78 & 37.93 & 38.72 & 68.84 & 38.14 & 49.08 & 39.08 & 41.90 \\\midrule
\textbf{Mistral-7B-v0.1} & 0 & 42.26 & 30.62 & 17.28 & 10.52 & 13.00 & 27.62 & 45.42 & 11.14 & 13.86 & 18.88 & 55.02 & 14.28 & 16.36 & 13.44 & 14.54 & 22.95 \\
 & 5 & 37.92 & 43.46 & 57.14 & 51.96 & 58.42 & 50.40 & 45.70 & 59.32 & 52.64 & 58.58 & 41.36 & 69.52 & 47.50 & 57.72 & 45.96 & 51.84 \\
 & 10 & 40.42 & 46.68 & 60.66 & 56.38 & 61.74 & 50.00 & 46.70 & 60.42 & 51.50 & 63.18 & 44.88 & 70.26 & 46.40 & 60.50 & 46.42 & 53.74 \\
 & 20 & 39.48 & 45.08 & 61.70 & 56.98 & 65.10 & 55.64 & 48.82 & 61.22 & 54.02 & 64.28 & 51.38 & 70.04 & 48.64 & 64.94 & 50.16 & 55.83 \\\midrule
\textbf{aya-23-35B} & 0 & 34.54 & 32.76 & 28.52 & 37.76 & 18.66 & 41.04 & 42.48 & 30.34 & 40.76 & 34.04 & 35.84 & 14.40 & 35.42 & 28.22 & 35.96 & 32.72 \\
 & 5 & 42.20 & 61.86 & 64.38 & 55.32 & 56.42 & 56.56 & 49.96 & 61.12 & 57.46 & 69.84 & 45.70 & 63.86 & 48.96 & 63.48 & 51.14 & 56.55 \\
 & 10 & 45.74 & 61.22 & 69.00 & 59.78 & 63.98 & 58.66 & 52.30 & 62.08 & 56.98 & 74.10 & 51.08 & 69.58 & 49.06 & 66.50 & 52.44 & 59.50 \\
 & 20 & 46.76 & 59.70 & 71.40 & 59.84 & 68.30 & 62.06 & 54.14 & 65.44 & 57.78 & 75.68 & 56.90 & 71.20 & 51.72 & 70.54 & 56.12 & 61.84 \\\midrule
\textbf{Llama-3.1-8B} & 0 & 29.14 & 34.10 & 36.04 & 55.82 & 22.60 & 48.60 & 42.96 & 39.84 & 55.00 & 49.04 & 23.18 & 15.20 & 46.18 & 38.02 & 47.42 & 38.88 \\
 & 5 & 41.50 & 50.12 & 63.90 & 46.56 & 64.48 & 42.24 & 34.94 & 61.22 & 39.08 & 63.88 & 35.80 & 68.12 & 43.66 & 58.14 & 43.30 & 50.46 \\
 & 10 & 45.16 & 51.70 & 66.04 & 47.72 & 66.12 & 44.60 & 39.16 & 62.70 & 36.90 & 68.54 & 39.88 & 69.84 & 43.20 & 60.88 & 42.68 & 52.34 \\
 & 20 & 50.20 & 50.56 & 67.92 & 48.18 & 66.92 & 48.50 & 44.10 & 64.28 & 38.70 & 72.00 & 48.26 & 70.40 & 42.72 & 62.24 & 43.16 & 54.54 \\\midrule
\textbf{Gemma-2-9B} & 0 & 48.26 & 37.78 & 26.88 & 17.60 & 16.24 & 34.34 & 46.86 & 26.68 & 16.22 & 37.12 & 57.98 & 27.20 & 18.20 & 22.96 & 16.58 & 30.06 \\
 & 5 & 61.10 & 63.62 & 65.72 & 55.20 & 62.84 & 63.24 & 56.34 & 63.36 & 54.46 & 74.78 & 62.12 & 65.44 & 46.24 & 70.14 & 52.12 & 61.11 \\
 & 10 & 61.26 & 63.78 & 65.98 & 60.64 & 66.38 & 64.60 & 56.40 & 64.08 & 57.28 & 78.16 & 61.20 & 69.76 & 49.66 & 73.02 & 54.74 & 63.13 \\
 & 20 & 64.14 & 61.86 & 67.90 & 60.26 & 69.38 & 65.48 & 57.24 & 64.50 & 58.62 & 79.24 & 62.98 & 70.68 & 50.02 & 73.96 & 56.00 & 64.15 \\\midrule
\textbf{Gemma-2-27B} & 0 & 53.40 & 46.38 & 44.10 & 33.62 & 31.08 & 49.02 & 50.82 & 54.20 & 38.16 & 59.88 & 55.58 & 31.34 & 29.66 & 45.96 & 31.90 & 43.67 \\
 & 5 & 61.40 & 66.00 & 69.40 & 60.44 & 68.30 & 63.26 & 56.34 & 64.44 & 62.34 & 80.18 & 59.90 & 64.34 & 51.36 & 75.60 & 56.24 & 63.97 \\
 & 10 & 62.44 & 64.76 & 69.74 & 64.70 & 70.62 & 66.22 & 56.24 & 64.22 & 61.24 & 81.94 & 59.52 & 70.82 & 53.18 & 75.34 & 59.16 & 65.34 \\
 & 20 & 64.12 & 63.60 & 70.24 & 65.74 & 72.84 & 67.72 & 59.18 & 66.06 & 63.20 & 82.98 & 64.44 & 73.28 & 56.04 & 76.50 & 61.18 & 67.14 \\\midrule
\textbf{Llama-3.1-70B} & 0 & 46.28 & 49.84 & 52.30 & 51.26 & 43.72 & 50.38 & 48.30 & 56.50 & 46.38 & 60.14 & 43.64 & 35.68 & 42.64 & 51.34 & 47.00 & 48.36 \\
 & 5 & 59.10 & 68.32 & 67.12 & 58.96 & 66.88 & 60.90 & 54.24 & 67.20 & 61.20 & 75.76 & 45.18 & 66.18 & 50.18 & 72.60 & 57.30 & 62.07 \\
 & 10 & 61.36 & 67.80 & 67.70 & 63.00 & 68.78 & 61.50 & 55.86 & 66.62 & 63.38 & 78.74 & 50.48 & 68.34 & 52.04 & 73.22 & 58.92 & 63.85 \\
 & 20 & 60.72 & 65.62 & 69.48 & 64.00 & 70.02 & 65.08 & 58.78 & 67.02 & 64.46 & 80.02 & 56.40 & 72.52 & 53.50 & 74.98 & 61.26 & 65.59 \\\midrule
\textbf{GPT-4o} & 0 & 61.70 & 56.05 & 66.45 & 45.90 & 44.21 & 44.92 & 44.93 & 49.67 & 38.50 & 57.24 & 46.01 & 33.90 & 36.72 & 48.62 & 36.34 & 47.41 \\
 & 5 & 66.69 & 73.26 & 71.03 & 53.24 & 74.23 & 72.39 & 65.30 & 55.93 & 56.19 & 79.02 & 57.64 & 61.36 & 51.81 & 69.65 & 58.45 & 64.41 \\
 & 10 & 67.09 & 75.12 & 73.81 & 55.28 & 78.02 & 74.47 & 66.97 & 56.47 & 58.34 & 72.83 & 56.50 & 61.11 & 61.61 & 70.78 & 65.12 & 66.23 \\
 & 20 & 67.22 & 76.05 & 75.00 & 56.08 & 78.17 & 74.85 & 66.01 & 50.33 & 61.95 & 84.14 & 56.91 & 51.67 & 62.78 & 72.22 & 61.84 & 66.35 \\\midrule\midrule
\textbf{Lang. avg.} & - & 47.21 & 47.15 & 51.73 & 45.76 & 51.10 & 48.39 & 47.03 & 49.50 & 44.74 & 56.22 & 47.65 & 50.37 & 42.15 & 51.59 & 43.85 & 48.30 \\
\bottomrule
\end{tabular}
}
\caption{Model Performances (Accuracy) for Zero and Few shot Learning}
\label{tab:llm_performance_acc}
\end{table*}

\begin{figure*}
    \centering
    \includegraphics[clip, trim=185 0 185 0, width=\textwidth]{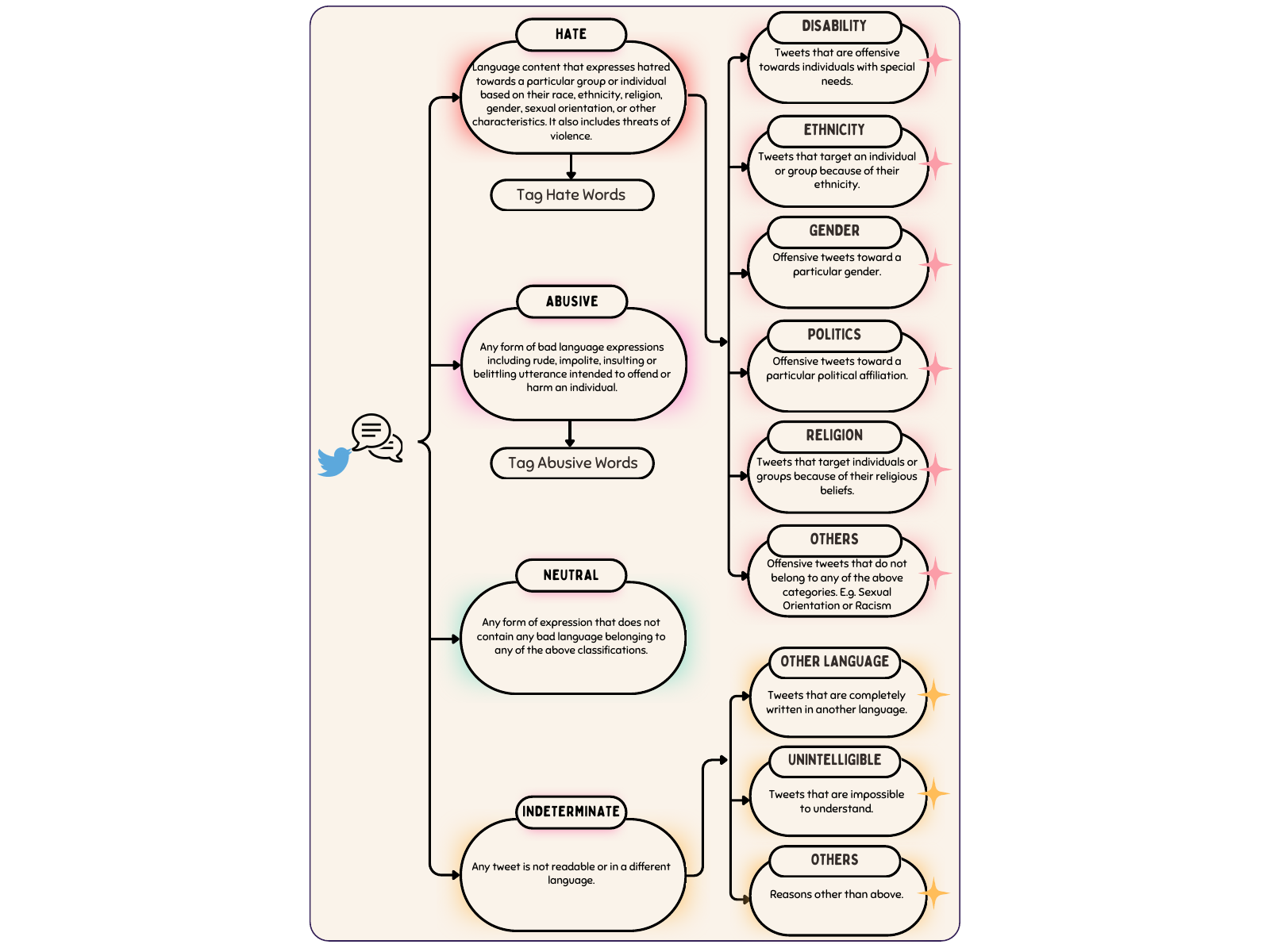}
    \caption{Annotation Guidelines and Definitions}
    \label{fig:annotation_guide}
\end{figure*}

\end{document}